\documentclass[10pt,twocolumn,letterpaper]{article}

\usepackage{cvpr}              
\usepackage{wrapfig}
\usepackage{multirow}
\usepackage[normalem]{ulem}
\useunder{\uline}{\ul}{}
\usepackage{array}

\usepackage[dvipsnames]{xcolor}

\definecolor{cvprblue}{rgb}{0.21,0.49,0.74}
\usepackage[pagebackref,breaklinks,colorlinks,citecolor=cvprblue]{hyperref}
\usepackage{graphicx}
\usepackage{multirow}

\def\paperID{16337} 
\def\confName{CVPR}
\def\confYear{2024}

\title{
OmniVec2 - A Novel Transformer based Network for Large Scale \\ Multimodal and Multitask Learning 
}

\author{Siddharth Srivastava, Gaurav Sharma\\
Typeface\\
{\tt\small \{siddharth.srivastava, gaurav\}@typeface.ai}
}

\begin{document}
\maketitle
\begin{abstract}

We present a novel multimodal multitask network and associated training algorithm.
The method is capable of ingesting data from approximately 12 different modalities
namely image, video, audio, text, depth, point cloud, time series, tabular, graph, X-ray, infrared, IMU, and hyperspectral.
The proposed approach utilizes modality specialized tokenizers, a shared transformer architecture, and cross-attention mechanisms to project the data from different modalities into a unified embedding space. It addresses multimodal and multitask scenarios by incorporating modality-specific task heads for different tasks in respective modalities. We propose a novel pretraining strategy with iterative modality switching to initialize the network, and a training algorithm which trades off fully joint training over all modalities, with training on pairs of modalities at a time. We provide comprehensive evaluation across 25 datasets from 12 modalities and show state of the art performances, demonstrating the effectiveness of the proposed architecture, pretraining strategy and adapted multitask training. 
\end{abstract}    
\section{Introduction}
\label{sec:intro}

Extracting meaningful representations from data is a central task in machine learning. Majority of the approaches proposed are usually specialized for specific modalities and tasks. The development of methods capable of handling multiple modalities, in a holistic way, has been an active topic of research recently~\cite{zhu2022uni,recasens2023zorro, girdhar2022omnivore,li2023uni, jaegle2021perceiver, jaegle2021perceiverio}. Multi task learning has a large body of literature \cite{mtlsurvey2020}, but has been traditionally limited to tasks from single modality. Learning a unified network that trains shared parameters across diverse tasks in different modalities, like image, video, depth maps, audio, has been shown to be more robust and give better generalization and reduce overfitting to a single task or modality \cite{akbari2021vatt, gong2022uavm} cf.\ unimodal networks. Such joint learning also enables more efficient use of available labeled data across various 
modalities, potentially reducing the need for extensive labeling in specific modalities for particular tasks. 

In the present work, we  extend such line of research and propose a multimodal multitask method which learns embeddings in a shared space across different modalities and then employs task specific sub-networks for solving specific tasks in specific modalities. 

The method utilizes a common transformer based bottleneck block to map the input to embeddings in a shared space, thus incorporating knowledge from multiple tasks associated with different respective modalities. This structure leads to learning of very robust representations informed and regularized by all tasks and modalities together. The embeddings are then used by the task heads to make required predictions.

Previous research in generalized multimodal learning falls into three main categories. First, there are methods that process multiple heterogeneous modalities such as images, 3D, and audio, directly without using separate encoders for each modality, learning representations directly from these inputs~\cite{jaegle2021perceiverio, jaegle2021perceiver}. Second, some approaches use modality specific encoders and then learn generalized embeddings, for data from each modality, based on a unified objective in the latent space~\cite{baevski2022data2vec}. Third, there are methods focused on knowledge sharing across different modalities, employing either a single common encoder~\cite{girdhar2022omnivore} or distinct encoders for each modality~\cite{akbari2021vatt}. Our work aligns more closely with the third type of approaches, while incorporating elements from the first. We employ modality specific tokenizers and encoders, and have a bottleneck shared transformer backbone. Tokenization is tailored to each modality, drawing inspiration from the Uni-Perceiver model but with key modifications detailed in Sec.~\ref{sec:approach}. After tokenization, transformer based network is used to obtain initial representations for the modalities which are passed through fully connected layers and then fused together with cross attention module. The fused representation then passes through the transformer backbone. The features from the transformer are then individually fused with original modality features using cross attention and are in turn fed to the modality specific task head. 

\begin{figure*}
    \centering
    \includegraphics[width=0.95\textwidth]{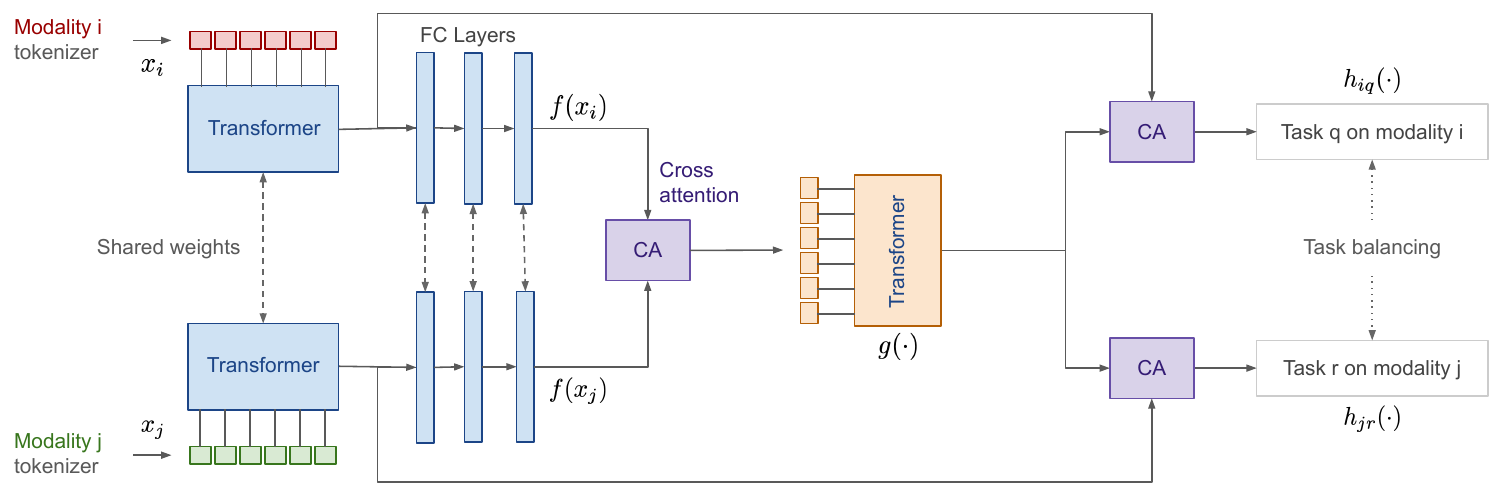}
    \caption{\textbf{Overview of the proposed method.} The proposed method consists of three parts, the feature transformation network $f(\cdot)$ which consists of a transformer followed by fully connected layers to reduce feature dimensions, another transformer $g(\cdot)$ and finally the task prediction heads $h_{mt}(\cdot)$ for task $t$ on modality $m$. The input data is tokenized with corresponding modality specific tokenizer. While training, pairs of modalities are used and the features are fused between the two modalities using cross attention layers, in a two stream configuration as shown here. While making prediction, the network is a single stream with cross attention layers removed, and the output is $h_{mt} \circ g \circ f (x)$ where $x$ is the output of the corresponding modality specific tokenizer.}
    \label{fig:arch}
\end{figure*}

The training procedure involves a dual-stage masked pretraining and a full task based loss optimization. The first stage of masked pretraining is the standard unsupervised masked pre-training with one modality at a time. The second state masked pretraining involves masked pretraining with pairs of modalities at a time, employing a two stream setup as shown in Fig.~\ref{fig:arch}. In this stage two modalities are used together, tokens are randomly masked and the full network is used to predict the masked tokens using the unmasked tokens for both modalities together. This allows for knowledge sharing across all modalities as the training proceeds by randomly sampling training batches from two modalities from all modalities. The final training step is then training for multiple tasks for different modalities. This is done similar to the second stage of masked pretraining, i.e.\ pairs of modalities are sampled, and a pair of tasks are sampled, one from each modality. Training batches are then constructed, half each from the two modality-task pairs. These are then used to optimize standard losses corresponding to the tasks, e.g.\ cross entropy for classification and $\ell_2$ loss for pixelwise prediction. The pretraining and final task training using pairs of modalities is the key component of the training strategy, that enables the cross modal knowledge sharing across all modalities together, which we discuss more in the following. 

In summary, the contributions of the work are as follows. (i) We propose a multimodal multitask network based on transformer architectures with modality specific tokenizers, shared backbone, and task specific heads. (ii) We provide comprehensive empirical results on 25 benchmark datasets over 12 distinct modalities \ie text, image, point cloud, audio and video along with applications to X-Ray, infrared, hyperspectral, IMU, graph, tabular, and time-series data. The method achieves better or close to state of the art performances on these datasets. (iii) We propose a novel multimodal pretraining approach that alternates between a pair of modalities to enable crossmodal knowledge sharing. (iv) We propose a multimodal and multitask supervised training approach to leverage knowledge sharing between modalities for robust learning, simplifying the complex processes proposed in previous works on modality integration, e.g.~\cite{yu2022metaformer, li2023uni}.
\section{Related Works}
\label{sec:relatedworks}

In this section, we discuss similar works and various similar paradigms to our work. 

\vspace{0.5em} 
\noindent\textbf{Multi-modal methods.} Contemporary multi-modal methods predominantly employ modality-specific feature encoders \cite{jiang2021review, kaiser2017one, arandjelovic2018objects, xiao2020audiovisual, recasens2023zorro}, focusing on fusion techniques within their architectural designs. These networks usually vary across modalities, necessitating architectural modifications for combined usage. They must address challenges related to feature fusion timing, fine-tuning, and pre-training \etc~\cite{xu2022multimodal}. Such complexities restrict the adaptability of universal frameworks like transformers for diverse domains, including point clouds, audio, and images. 
\vspace{0.5em} 

\noindent\textbf{Common network for multiple modalities.} A growing body of research aims to learn from multiple modalities without modality-specific encoders~\cite{ carreira2022hierarchical, girdhar2022omnivore, baevski2022data2vec, jaegle2021perceiver}. Notably, architectures like the perceiver~\cite{jaegle2021perceiver, jaegle2021perceiverio, carreira2022hierarchical} employ cross-attention among latent queries to process multiple modalities together. The hierarchical perceiver~\cite{carreira2022hierarchical} expands on this by structuring the input while maintaining locality. Other approaches, such as data2vec~\cite{baevski2022data2vec}, use modality-specific encoders. Omnivore~\cite{girdhar2022omnivore}, with a common encoder, is limited to visual modalities only. Contrarily, VATT~\cite{akbari2021vatt} employs a unified transformer backbone but processes each modality independently. These multi-modal methods have demonstrated enhanced robustness \cite{akbari2021vatt, gong2022uavm}.
\vspace{0.5em} 

\noindent\textbf{Multi-task learning.} As explored in the preceding section, there has been a surge in methods that process multiple modalities. PerceiverIO\cite{jaegle2021perceiverio} extends the capabilities of Perceiver~\cite{jaegle2021perceiver} to facilitate learning multiple tasks with a singular network architecture. Although PerceiverIO is capable of multitasking, often separate networks are employed~\cite{zhang2018overview}. Various techniques~\cite{baevski2022data2vec, girdhar2022omnivore, hu2021unit, pramanik2019omninet, dai2022one} learn from raw representations of multiple modalities and are applicable to numerous tasks.
\vspace{0.5em} 

\noindent\textbf{Multi-modal masked pretraining.}
Approaches such as \cite{liu2021opt, yan2022multi, wei2022masked} implement masked pre-training. This technique has proven beneficial for improving the performance of deep networks across different modalities and tasks\cite{akbari2021vatt, girdhar2022omnimae, baevski2022data2vec, baade2022mae, yu2022point, gupta2022maskvit}.

\vspace{0.5em} 

\noindent\textbf{Comparison to similar works.} We draw motivations from UniPerceiver~\cite{zhu2022uni}, MetaFormer~\cite{zhang2023meta} and OmniVec~\cite{srivastava2023omnivec}. Unlike UniPerceiver line of methods, we do not use a unified task head definition, while similar to it we use task specific task heads. This allows our method to learn more robust and leverage fine details from each task depending upon the complexity of the tasks, which is important as each modality has distinct definition of complexity. For ex., in vision task, classification is a relatively simpler task as compared to segmentation, as segmentation tasks enforces networks to learn pixel level attention and learning better neighbourhood relationships~\cite{guo2018dynamic, srivastava2023hierarchical}. Further,  MetaFormer uses unified tokenizers, and instead, we utilize modality specific tokenizers. Our experiments indicate that modality specific tokenizers perform better than MetaFormer's unified tokenizer when training on multiple modalities. Further, OmniVec uses separate encoders for each modaity, that makes the network heavy and computationally expensive. In contrast, we use modality specific tokenizers with a shared backbone. Additionally, unlike other works, we train on multiple modalities in a multi task manner, allowing the network to learn from multiple modalities with varying task complexities simultaneously.

\section{Approach}
\label{sec:approach}

\paragraph{Overview.}
We are interested in multimodal multitask learning. Say we have modalities indexed by $m \in [1,M]$, and each modality has $T$ tasks indexed by $t \in [1,T]$. Note that here we assume same number of tasks for all modalities for notational convenience, in practice different modalities would have different number of tasks. Examples of modality and their tasks could be classification into categories for point cloud modality, and dense pixel wise segmentation in image modality. We are interested in jointly learning classifiers $\phi_{mt}(\cdot | \theta_{mt})$ which take inputs $x_m$ from modality $m$ and make predictions for task $t$, with $\theta_{mt}$ being the respective parameters. We assume that the learning is to happen by loss minimization where $\ell_{mt}(\cdot)$ denotes the loss for task $t$ on modality $m$. Examples of such losses are cross entropy loss for classification tasks, and $\ell_2$ loss for dense image prediction tasks such as image segmentation. We would like to solve the following optimization.

\begin{equation}
\label{eqn:obj}
\Theta^* = \min_{ \Theta } \sum_{m,t} \ell_{mt}(\mathcal{T}_{mt}),
\end{equation}
where $\Theta = \{ \theta_{mt} | m, t\}$ are the parameters of all the predictors, and $\mathcal{T}_{mt}$ is the training set provided for task $t$ of modality $m$. This is the extension of multiple task learning to multiple modalities as well. 

We present a network and associated unsupervised pretraining and supervised training algorithm for the above task of multimodal multitask learning. The network consists of  $M\times T$ modality specific tokenizers, followed by common feature transformation and feature fusion networks built with transformers, with cross attention modules in between, denoted by $f(\cdot), g(\cdot)$ in Fig.~\ref{fig:arch}. The final part of the network are $M\times T$ task specific prediction heads, denoted by $h_{mt}(\cdot)$ for task $t$ on modality $m$, which provide the final outputs for the tasks. At inference the prediction function is the composition of the three functions, i.e.\ $\phi(x) = h_{mt} \circ g \circ f (x)$ where $x$ is the tokenized form of the input.  While training, we sample a pair of modalities from all the available modalities, and then sample one task each for the sampled modalities. We then construct training batch, half from each sampled task. Once the tokenization is done, the features $x_i, x_j$ are passed into the first feature transformation subnetwork to obtain $f(x_i), f(x_j)$. These are then passed through the cross attention module to fuse them together. The fused features are then input to the second part of the network, i.e.~$g(\cdot)$. The output $\hat{x}_{ij} = g \circ \mathcal{A} ( f(x_i), f(x_j))$, where $\mathcal{A}(\cdot)$ is the cross attention function, is then again fused with the respective input features $x_i, x_j$. These features, i.e. $\mathcal{A}( \hat{x}_{ij}, x_i), \mathcal{A}( \hat{x}_{ij}, x_j)$ are then fed to the task predictors $h_{iq}$ and $h_{jr}$, to obtain the final predictions for task $q,r$ on modalities $i,j$ respectively. The sum of losses $\ell_{iq} + \ell_{jr}$ are then minimized for the current batch by backpropagation. Thus the learning proceeds by optimizing pairs of losses at a time, to stochastically minimize the sum over all the losses.

Along with the supervised multimodal joint training explained above, the learning also consists of two stages of unsupervised masked pretraining with the first stage being unimodal and the second stage being multimodal pretraining, to achieve knowledge sharing between tasks and modalities leading to regularized and robust predictors. We now present each of the components and the full training algorithm in detail.

\subsection{Network components} 

We now go through the network components sequentially from input to output.
\vspace{0.5em} \\
\noindent\textbf{Tokenizers.} Each modality is tokenized using a modality specific tokenizer. The tokenizers are similar to those used in Uni-Perceiver~\cite{li2023uni}, however, instead of attaching an embedding to the tokens, we provide transformer with one type of modality at a time. Further, Uni-Perceiver utilizes a combination of tokens from multiple modalities passed to a single transformer. This limits the Uni-perceiver to a limited set of modalities, i.e.\ text, image and videos. However, our method does not suffer from any such limitation. The details of specific tokenizers for the different modalities are provided in Supplementary.  
\vspace{0.5em} \\
\noindent\textbf{Feature transformation network.} Once the features are tokenized, they are then passed through a transformer network. While the method can utilize any transformer backbone, in the current implementation we use a transformer based on BERT~\cite{devlin2018bert}. Here, the multi head attention involves standard self-attention~\cite{vaswani2017attention}, and GeLU~\cite{hendrycks2016gaussian} activation prior to the MLP layer. The output from the transformer network is passed to a fully connected neural network with three fully connected layers with ReLU activation. This transformer network along with the fully connected layers is denoted a $f(\cdot)$ in Fig.~\ref{fig:arch}. The network could be used without the fully connected layers---we added the fully connected layers to reduce the dimensions of the features so that the computational complexity of the remaining part of the network could be reduced.
\vspace{0.5em} \\
\noindent\textbf{Mixing features with cross attention.} When training, we fuse the features from the two transformer streams, corresponding to two modalities, with cross attention module. The output fused features are then passed to another transformer network, denoted a $g(\cdot)$ in Fig.~\ref{fig:arch}. The architecture of the transformer network is same as the transformers used in feature transformation network.
\vspace{0.5em} \\
\noindent\textbf{Modality and task specific heads.} The part of the network are the modality and task specific heads, denoted a $h_{mt}(\cdot)$ in Fig.~\ref{fig:arch}. These task heads take as input, features from respective modality streams fused with features from the above network, fused with cross attention module. The task heads consist of a vanilla ViT-Tiny networks~\cite{wu2022tinyvit}.

\vspace{-0.5em}
\subsection{Training}\label{subsec:training}

The training is done in three steps: (i) masked pretraining iterating over modalities but doing masked prediction with one modality at a time, (ii) multimodal masked pretraining where two modalities are simultaneously used to do masked prediction for each, and (iii) finally supervised task based training.
\vspace{0.5em} \\
\noindent\textbf{Stage 1 masked pretraining.} The first step in training is self supervised pretraining of the transformer in the feature transformation network.  We follow earlier works~\cite{akbari2021vatt, srivastava2023omnivec, girdhar2022omnimae} and add a decoder for predicting masked tokens. Specifically, for an input modality with $P$ patches, we randomly mask $P_m$ patches, and feed non-masked patches and their positions to an encoder network attached in addition to the feature transformer. Further, we iterate between modalities while keeping the transformer network common, so that it learns to work with all modalities. Once this stage is complete we discard the decoder added, and keep only the encoder transformer.
\vspace{0.5em} \\
\noindent\textbf{Stage 2 masked pretraining.} We engage the full network, except the task specific prediction heads. We take two inputs from two different modalities and pass them through the network till just before the task prediction heads. Instead of task prediction heads we add decoders to predict the masked tokens for respective input modalities. This process involves decoding the modalities in parallel, utilizing the outputs from the cross-attention modules and the modality-specific feature vectors. This alternating approach is key to achieving effective multimodal masked pretraining. Here also, we randomly mask tokens for both the modalities. Task balancing is not employed in this pretraining stage. Such a multi task multi modality approach allows us to utilize unpaired data across modalities. As in stage 1 pretraining, once this stage of training is finished, we discard the decoders added and keep the trained network $f, g$.

\vspace{-0.8em}
\subsubsection{Multimodal multitask supervised training}
In the final part of the training, we train for two tasks at a time from two different modalities. This lets up stochastically minimize the loss function in Eq.~\ref{eqn:obj}, but minimizing sum of two losses at a time instead of minimizing the sum of all of them. When we use two modalities, we use the network as shown in Fig.~\ref{fig:arch} in a two stream configuration. With the two modality features being fused together in the middle, passed through a transformer $g(\cdot)$ and then fused back with themselves, before finally being input to the task prediction heads. Such fusion of the the features from two modalities leads to knowledge sharing between the tasks of different modalies and makes the learning robust and regularized.

Given the varying complexities of these task pairs, as underscored in previous research~\cite{fifty2021efficiently}, we found it essential to balance the complexity of tasks in a multitask learning setting. Hence, the we train while employing standard task balancing techniques. We adjust the loss magnitude for each task based on its convergence rate. As our ablation studies will demonstrate, this approach allows for random pairing of modalities, in contrast to the need for selecting specific pairs as suggested in prior works~\cite{srivastava2023omnivec, yu2022metaformer,zhu2022uni, li2023uni}. We give details of such task balancing in the Supplementary material.

\vspace{-0.5em}
\subsubsection{Masked pretraining for different modalities} 

We use the best practices when pretraining with different modalities, following existing works. We use image, video, text, audio and 3D point clouds modalities for masked pretraining. We employ a consistent masking approach across visual and auditory modalities. We follow~\cite{song2020mpnet} for textual data, utilizing random sentence permutation~\cite{yang2021generalized}. We designate a fraction $f$ of tokens for prediction, following the 8:1:1 token masking ratio of BERT~\cite{devlin2018bert}. Our primary goal is to reduce the discrepancy between the input and the outputs of the decoder. For inputs such as images, videos, point clouds, and audio spectrograms, we aim to minimize the $\ell_2$ distance between the predicted and actual target patches. Normalization to zero mean and unit variance is applied to visual inputs. For textual data, we utilize the permuted language modeling objective of XLNet~\cite{yang2021generalized}.

\vspace{-0.5em}
\subsubsection{Inference}

When doing prediction, the network is used as a single stream without the cross attention layers in Fig.~\ref{fig:arch}. The input data is tokenized with the tokenizer for its modality, passed through the feature transformation network $f(\cdot)$ followed by the second transformer $g(\cdot)$, and finally input to the task prediction head $h_{mt}(\cdot)$, i.e.\ the full forward pass is $h_{mt} \circ g \circ f (x)$ where $x$ is the output of the tokenizer.
\section{Experimental results}

\begin{table*}
\centering
\parbox{0.3\textwidth}{
\centering
\resizebox{0.3\textwidth}{!}{ 

\begin{tabular}{lm{3.6em}m{3em}}
\hline
\textbf{Method/Dataset}         & \textbf{iN2018} & \textbf{P365}                  \\
\hline
Omni-MAE~\cite{girdhar2022omnimae}                       & 78.1            & 59.4                           \\
Omnivore~\cite{girdhar2022omnivore}                        & 84.1            & 59.9                           \\
EfficientNet B8\cite{tan2019efficientnet}                 & 81.3            & 58.6                           \\
MAE\cite{he2022masked}                             & 86.8            &                                \\
MetaFormer~\cite{yu2022metaformer}                      & 87.5            & 60.7                           \\
InternImage\cite{wang2023internimage} & 92.6   & 61.2 \\
OmniVec~\cite{srivastava2023omnivec}                         & {\ul 93.8}      & {\ul 63.5}  \\

Ours                         & \textbf{94.6}      & \textbf{65.1}  \\
\hline
\end{tabular}}
\caption{\textbf{iNaturalist-2018 and Places-365} top-$1$ accuracy. 
}
\label{tab:sota_image}
}
\hspace{0.5em}
\parbox{0.45\columnwidth}{
\centering
\resizebox{0.45\columnwidth}{!}{ 

\begin{tabular}{lr}
\hline
\textbf{Method} & \textbf{K400} \\
\hline
Omnivore~\cite{girdhar2022omnivore}                & 84.1                  \\
VATT~\cite{akbari2021vatt}                    & 82.1                  \\
Uniformerv2~\cite{li2022uniformerv2} & 90.0 \\
InternVideo\cite{wang2022internvideo}             & {\ul 91.1}         \\
TubeViT\cite{piergiovanni2023rethinking}                 & 90.9                  \\
OmniVec\cite{srivastava2023omnivec}                 & {\ul 91.1} \\     

Ours                 & \textbf{93.6} \\\hline                

\end{tabular}}
\caption{\textbf{Kinetics-400} top-$1$ accuracy.
}
\label{tab:sota_kinects400}
}
\hspace{0.5em}
\parbox{0.45\columnwidth}{
\centering
\resizebox{0.45\columnwidth}{!}{ 

\begin{tabular}{lm{4em}}
\hline
\textbf{Method} & \textbf{MIT} \\
\hline
VATT~\cite{akbari2021vatt}                    & 41.1                     \\
Uniformer v2\cite{li2022uniformerv2}           & 47.8                     \\
CoCa\cite{yu2022coca}                    & 47.4                     \\
CoCa-finetuned\cite{yu2022coca}          & 49.0               \\
OmniVec\cite{srivastava2023omnivec}                 & {\ul 49.8} \\

Ours                 & \textbf{53.1} \\
\hline
\end{tabular}
}
\caption{\textbf{Moments in time} top-$1$ accuracy.
}
\label{tab:sota_MIT}
}
\hspace{0.5em}
\parbox{0.4\columnwidth}{
\resizebox{0.4\columnwidth}{!}{
    \begin{tabular}{lr}
        \hline
        \textbf{Method} & \textbf{ESC50} \\
        \hline
        AST~\cite{gong2021ast}                     & 85.7           \\
        EAT-M\cite{gazneli2022end}                   & 96.3           \\
        HTS-AT\cite{chen2022hts}                  & 97.0           \\
        BEATs\cite{oreshkin2019n}                   & 98.1     \\
        OmniVec\cite{srivastava2023omnivec}                 & {\ul 98.4} \\
        
        Ours                 & \textbf{99.1} \\
        \hline
    \end{tabular}
}
\caption{\textbf{ESC50} top-$1$ accuracy. 
}
\label{tab:sota_audio}
}
\hspace{1em}
\parbox{0.45\columnwidth}{
\resizebox{0.45\columnwidth}{!}{

}
}
\end{table*}
\begin{table*}
\centering
\parbox{0.6\columnwidth}{
\centering
\resizebox{0.55\columnwidth}{!}{
\begin{tabular}{lm{4em}}
\hline
\textbf{Method} & \textbf{MN40C} \\
\hline
PointNet++\cite{qi2017pointnet++}              & 0.236                 \\
DGCN+PCM-R\cite{zhang2022pointcutmix}     & 0.173                 \\
PCT + RSMIx\cite{lee2021regularization}             & 0.173                 \\
PCT + PCM-R\cite{sun2022benchmarking}     & 0.163           \\
OmniVec\cite{srivastava2023omnivec}                 & {\ul 0.156}   \\  

Ours                 & \textbf{0.142}   \\ \hline 
\end{tabular}
}
\caption{\textbf{ModelNet40-C} Error Rate.
}
\label{tab:sota_mnc}
}
\hspace{1em}
\parbox{0.6\columnwidth}{
\centering
\resizebox{0.6\columnwidth}{!}{
\begin{tabular}{lr}
\hline
\textbf{Method} & \textbf{S3DIS} \\
\hline
PointTransformer+CBL\cite{tang2022contrastive}    & 71.6           \\
StratifiedTransformer\cite{lai2022stratified}   & 72.0           \\
PTv2\cite{wu2022point}                    & { 72.6}     \\
Swin3D\cite{yang2023swin3d}                   & 74.5     \\
OmniVec\cite{srivastava2023omnivec}                 & {\ul 75.9}  \\

Ours                 & \textbf{77.1}  \\
\hline
\end{tabular}
}
\caption{\textbf{Stanford Indoor Dataset} mIoU.
}
\label{tab:sota_s3dis}
}
\hspace{1em}
\parbox{0.75\columnwidth}{
\centering
\resizebox{0.7\columnwidth}{!}{
\begin{tabular}{ccccc}
\hline
\textbf{Method} & \textbf{R-1}   & \textbf{R-2}   & \textbf{R-L}   & \textbf{B-S}  \\
\hline
CODS\cite{wu2021controllable}            & 44.27          & 17.90          & 36.98          & 70.49         \\
SICK\cite{kim2022mind}            & 46.2     & 20.39    & {\ul 40.83} & 71.32   \\
OmniVec\cite{srivastava2023omnivec}         & {\ul 46.91} & {\ul 21.22} & 40.19    & {\ul 71.91} \\

Ours         & \textbf{47.6} & \textbf{22.1} & \textbf{41.4}    & \textbf{72.8} \\
\hline
\end{tabular}
}
\caption{\textbf{DialogueSUM} text summarization ROGUE scores. 
}
\label{tab:dialoguesum}
}
\end{table*}

\noindent\textbf{Masked pretraining.} We use AudioSet (audio) \cite{gemmeke2017audio}, Something-Something v2 (SSv2) (video) \cite{goyal2017something}, English Wikipedia (text), ImageNet1K (image) \cite{deng2009imagenet}, SUN RGB-D (depth maps) \cite{song2015sun}, ModelNet40 (3D point cloud) \cite{wu20153d} for pretraining the network. For Stage 1 of masked pretraining (Sec.~\ref{subsec:training}), we use the samples from the training set of the respective datasets. For Stage 2 of masked pretraining, we randomly select two modalities, and sample data from them to pretrain the full network. 
Further, we randomly mask patches. For image, video and audio, we mask $95\%$ of the patches. For point cloud and text, we mask $90\%$ and $95\%$ of the patches respectively. We perform pretraining for $3000$ epochs. We use fraction $f$ as $5\%$.
\vspace{0.5em} \\
\noindent\textbf{Downstream tasks.} We train the model on downstream tasks and report results. The datasets used for single modality methods are iNaturalist-2018 \cite{van2018inaturalist} (Image Recognition), Places-365 \cite{zhou2017places} (Scene Recognition), Kinetics-400 \cite{kay2017kinetics} (Video Action Recognition), Moments in Time \cite{monfort2019moments} (Video Action Recognition), ESC50 \cite{piczak2015esc} (Audio Event Classification), S3DIS \cite{armeni20163d} (3D point cloud segmentation), DialogueSUM \cite{chen2021dialogsum} (Text summarization).
\vspace{0.5em} \\
\noindent\textbf{Adaptation on unseen datasets.} To assess our method's adaptability to datasets not seen at training, we report comparisons with image classification on Oxford-IIIT Pets \cite{parkhi2012cats}, action recognition in videos using UCF-101 \cite{soomro2012ucf101} and HMDB51 \cite{kuehne2011hmdb}, 3D point cloud classification on ScanObjectNN \cite{uy2019revisiting}, point cloud segmentation with NYU v2 seg \cite{silberman2012indoor}, text summarization using the SamSum dataset \cite{gliwa2019samsum}. 
As the number of classes and labels differ in each dataset as compared to the datasets used during pretraining, we randomly sample $10\%$ data from each of the training set. Further, we extract the embeddings using the pretrained network, and train two fully connected layers with task specific loss functions. This allows us to demonstrate the ability of the proposed method to generate embeddings which can generalize across datasets.
\vspace{0.5em} \\
\noindent\textbf{Cross domain generalization.} We follow prior work~\cite{akbari2021vatt} and evaluate on video-text retrieval on two benchmark datasets \ie YouCook2~\cite{zhou2018towards}, and MSR-VTT~\cite{xu2016msr}, for multiple modalities. 
\vspace{0.5em} \\
\noindent\textbf{Adaptation on unseen modalities.} We also evaluate our method on unseen modalities. Specifically, we evaluate our method on the following
(i) X-Ray scan, and hyperspectral data recognition, where we utilize the RegDB~\cite{nguyen2017person}, Chest X-Ray~\cite{rahman2020reliable}, and Indian Pine datasets\footnote{\url{https://github.com/danfenghong/IEEE_TGRS_SpectralFormer/blob/main/data/IndianPine.mat}}. (ii) Time-series forecasting, where our experiments are based on the ETTh1~\cite{haoyietal-informer-2021}, Traffic\footnote{\url{https://pems.dot.ca.gov/}}, Weather\footnote{\url{https://www.bgc-jena.mpg.de/wetter/}}, and Exchange datasets~\cite{lai2018modeling}. (iii) Graph understanding through the PCQM4M-LSC dataset~\cite{hu2021ogb}, which comprises 4.4 million organic molecules with quantum-mechanical properties, focusing on predicting molecular properties with applications in drug discovery and material science. (iv)Tabular analysis, where we engage with the adult and bank marketing datasets from the UCI repository\footnote{\url{http://archive.ics.uci.edu/ml/}}, (v) IMU recognition, where we conduct experiments on IMU sensor classification using the Ego4D dataset~\cite{grauman2022ego4d}, assessing the capability to understand inertial motion systems. 
We follow~\cite{zhang2023meta} for the train test splits and evaluation metrics on these datasets. Further, we use modality specific tokenizers and follow similar network settings as for generalization on unseen datasets. 

\vspace{0.5em} 

\noindent We provide more details on the tokenizers used for each modality, description of task heads, and formulations of loss functions in the supplementary material.

\begin{table*}
\centering
\resizebox{0.95\textwidth}{!}{ 
\begin{tabular}{llllcll}
\hline
\textbf{Dataset} & \textbf{Modality} & \textbf{Task} & \textbf{Metric}  & \textbf{Ours}   & \textbf{SOTA}  & \textbf{Ref.}       \\
\hline
UCF-101 &  Video & Action Recognition  &   3-Fold Accuracy  & {\ul 99.1}   &    \textbf{99.6} & OmniVec~\cite{srivastava2023omnivec} \\
HMDB51 & Video  & Action Recognition  &   3-Fold Accuracy  & \textbf{92.1}   &   {\ul 91.6} & OmniVec~\cite{srivastava2023omnivec} \\
Oxford-IIIT Pets &  Image & Fine grained classification  &   Top-1 Accuracy  & \textbf{99.6} &   \uline{99.2} & OmniVec~\cite{srivastava2023omnivec} \\
ScanObjectNN   & 3D Point Cloud & Classification  &   Accuracy  & \textbf{97.2}  &   \uline{96.1} & OmniVec~\cite{srivastava2023omnivec} \\
NYU V2 & RGBD  & Semantic Segmentation  &   Mean IoU  & \textbf{63.6} &   \uline{60.8} & OmniVec~\cite{srivastava2023omnivec} \\
SamSum & Text  & Meeting Summarization  &   ROGUE(R-L)  & \textbf{55.4} &   \uline{54.6} & OmniVec~\cite{srivastava2023omnivec} \\
YouCook2 & Video+Text  & Zero Shot Text-to-Video Retrieval  &   Recall@10  & \uline{69.9}  &   64.2(Pre) / \textbf{70.8}(FT) & OmniVec~\cite{srivastava2023omnivec} \\
MSR-VTT & Video+Text  & Zero Shot Text-to-Video retrieval  &   Recall@10  & \uline{85.8}  &  80.0(Pre) / \textbf{90.8}(FT) & SM \cite{zeng2022socratic} \\
\hline
\end{tabular}
}
\caption{Adaptation on \textit{unseen datasets.} (Oxford-IIIT Pets, UCF-101, HMDB51, ScanObjectNN, NYUv2 Seg, SamSum), and \textit{cross-domain} generalization (YouCook2, MSR-VTT). See supplementary for more detailed results.}
\label{tab:generalization}
\end{table*}

\subsection{Comparison with state of the art methods}
We performed masked pretraining followed by training on multiple modalities and task groups as described in Section \ref{sec:approach} for comparing with existing methods. We discuss the comparison on each modality below.
\vspace{0.5em} \\
\noindent\textbf{Image.}
Table~\ref{tab:sota_image} shows state of the art on iNaturalist 2018 and Places 365 datasets. On the iNaturalist 2018 dataset, our method achieves a top-1 accuracy of 94.6\%, surpassing notable contenders such as OmniVec (93.8\%), MetaFormer (87.5\%), and MAE (86.8\%). This superior accuracy demonstrates capability of the proposed method in accurately recognizing a diverse range of natural species. In the context of the Places 365 dataset, our method achieves an accuracy of 65.1\%, notably outperforming OmniVec (63.5\%), and significantly surpassing MetaFormer's 60.7\% and Omnivore's 59.9\%. The substantial margin of improvement, particularly in the challenging and variable environment of Places 365, underscores the robustness and adaptability of the proposed architecture.  We also conduct experiments on ImageNet~\cite{deng2009imagenet} (classification), MSCOCO~\cite{lin2014microsoft} (object detection), and ADE-20K~\cite{zhou2017scene} (semantic segmentation) datasets (detailed table is in supplementary).  89.3\% (accuracy) on ImageNet, 60.1 (AP) on MSCOCO and an mIoU of 58.5 on ADE-20K. 
\vspace{0.5em} \\
\noindent\textbf{Video.} Table~\ref{tab:sota_kinects400} and Table~\ref{tab:sota_MIT} show comparison against state of the art methods on Kinetics-400 and Moments in Time datasets.We observe that we outperform all the competing methods on both the datasets achieving top-1 accuracy of $93.6\%$ and $53.1\%$ respectively. 
\vspace{0.5em} \\
\noindent\textbf{Audio.} 
Table~\ref{tab:sota_audio} shows our comparison with top-performing methods on the ESC50 dataset. We outperform competing methods, achieving an accuracy of 99.1\%, significantly higher than the Audio Spectrogram Transformer (AST) at 85.7\%, and OmniVec at 98.4\%.
\vspace{0.5em} \\
\noindent\textbf{Point Cloud.} Table~\ref{tab:sota_mnc} and Table~\ref{tab:sota_s3dis} compare against state of the art methods on ModelNet40-C and S3DIS datasets respectively. On ModelNet40-C, we evaluate a classification task, while on S3DIS we evaluate semantic segmentation. On both the datasets, we outperform the competing methods. On ModelNet-C, we achieve an error rate of 0.142, which is notably lower than the rates observed in other contemporary methods. This is particularly evident when compared against methods like OmniVec, which recorded an error rate of 0.156, and PCT + PCM-R, with an error rate of 0.163. On S3DIS,  we achieve an mIoU of 77.1, which is the highest among all the methods evaluated c.f. 75.9 of  OmniVec, and 74.5 of Swin3D. This demonstrates that the proposed method is able to obtain a robust performance with the shared backbone network across tasks. 
\vspace{0.5em} \\
\noindent\textbf{Text.} 
Table \ref{tab:dialoguesum} shows state of the art on DialogueSUM dataset for text summarization. Our method surpasses other methods in all the metrics. Despite utilizing significantly fewer datasets for text in comparison to visual tasks , our method demonstrates strong performance. This suggests proposed method's capacity to bridge the modality gap~\cite{liang2022mind} across distinct domains in the latent space, even when the data distribution is skewed.

Table \ref{tab:glue} illustrates the experimental results on the GLUE benchmark for text understanding tasks, comparing various state-of-the-art methods such as BERT~\cite{devlin2018bert}, RoBERTa~\cite{liu2019roberta}, and ChatGPT. The comparison centers on paraphrasing, sentiment, duplication, inference, and answering tasks. We achieve second best performance on three out of five tasks demonstrating its capability to perform reasoning and adaptability to natural language tasks. 
\\
\noindent\textbf{Comparison on pretraining datasets.} We fine tune our pretrained network on the respective training sets with related task heads. We obtain an mAP of 55.8 and 56.4 on AudioSet(A) and AudioSet(A+V) respectively. Further, on SSv2, ImageNet-1K, SUN-RGBD, and ModelNet we achieve top-1 accuracies of 86.1\%, 93.6\%, 75.9\% and 97.1\% respectively. We outperform the competing state of the art methods on these datasets(detailed results are in supplementary). 

\begin{table}[]
\resizebox{1.0\columnwidth}{!}{
\begin{tabular}{l|ccccc}
\hline
\multirow{3}{*}{Method}                                                    & \multicolumn{5}{c}{GLUE Benchmark}                                            \\
                                                                           & SST-2         & MRPC          & QQP           & MNLI          & QNLI          \\
                                                                           & Sentiment     & Paraphrase    & Duplication   & Inference     & Answering     \\ \hline
BiLSTM+ELMo+Attn                                                           & 90.4          & 84.9          & 64.8          & 76.4          & 79.8          \\ \hline
OpenAI GPT~\cite{radford2018improving}               & 91.3          & 82.3          & 70.3          & 82.1          & 87.4          \\
$\text{BERT}_{\text{BASE}}$~\cite{devlin2018bert}    & 88.0          & {\ul 88.9}    & 71.2          & 84.6          & {\ul 90.5}    \\
$\text{RoBERTa}_{\text{BASE}}$~\cite{liu2019roberta} & \textbf{96.0} & \textbf{90.0} & \textbf{84.0} & 84.0          & \textbf{92.0} \\
$\text{ChatGPT}$                                                           & 92.0          & 66.0          & 78.0          & \textbf{89.3} & 84.0          \\ 
$\text{Meta-Transformer-B16}_\text{T}$~\cite{zhang2023meta}                                     & 81.3          & 81.8          & 78.0          & 70.0          & 60.3          \\ \hline
$\text{Ours}$                                                              & {\ul 95.6}    & 85.8          & {\ul 82.2}    & {\ul 87.9}    & 84.2          \\ \hline
\end{tabular}}
\caption{\textbf{Text understanding on the GLUE benchmark.} We compare existing advanced methods from paraphrasing, sentiment, duplication, inference, and answering tasks.}
\label{tab:glue}
\end{table}

\begin{table*}[h]
\parbox{0.67\linewidth}{
\centering

	\centering
	\vspace{2mm}
	\subfloat[SYSU-MM01 (infrared)~\label{tab:infrared}]{%
		\resizebox{0.47\linewidth}{!}{
			\begin{tabular}{l|c|c}
				\toprule
				\bfseries Method                 & \bfseries R@1 (\%)               & \bfseries {mAP} (\%)      \\ \midrule
				AGW \cite{arxiv20reidsurvey}        & 70.49           & {65.90}          \\
				SMCL~\cite{Wei_2021_ICCV} & 83.05 &{\uline{78.57}} \\
				MSCLNet~\cite{zhang2022modality} & \uline{83.86} & 78.31 \\
				$\text{Meta-Transformer-B16}_\text{F}$~\cite{zhang2023meta}  & {73.50}  & {65.19}  \\
    \hline
    $\text{Ours}$~  & \textbf{86.21}  & \textbf{84.24}  \\
				\bottomrule
			\end{tabular}
		}
	}
	\subfloat[Indian Pine (hyperspectral)~\label{tab:hyper}]{
		\resizebox{0.45\linewidth}{!}{
			\begin{tabular}{l|c|c}
				\toprule
				\bfseries Method &\bfseries OA (\%) & \bfseries AA (\%) \\ 
				\midrule
				ViT~\cite{dosovitskiy2020image} & 71.86 & 78.97   \\
				SpectralFormer~\cite{hong2021spectralformer} (Pixel)  & 78.55 & 84.68  \\
				SpectralFormer~\cite{hong2021spectralformer}(Patch) & 81.76 & \uline{87.81}  \\
                RPNet-RF~\cite{uchaev2023small} & \uline{90.23} & \\
                HyLITE~\cite{zhou2023locality} & 89.80 & \\
                TC-GAN~\cite{bai2022generative} & 87.47 & \\

				$\text{Meta-Transformer-B16}_\text{F}$~\cite{zhang2023meta} & 67.62 & 78.09  \\
                \hline
                $\text{Ours}$~ & \textbf{90.6} & \textbf{89.3} \\
				\bottomrule
			\end{tabular}
		}
	} 
 \caption{\textbf{Infrared and hyperspectral classification}. Metrics are Rank-1 (R@1), mean Average Precision (mAP), Overall Accuracy (OA), Average Accuracy (AA).
 }
 \label{tab:ablation}
 }
 \hspace{1em}
 \parbox{0.29\linewidth}{
\centering
\resizebox{0.98\linewidth}{!}{
	\begin{tabular}{l|p{2em}p{2em}}
		\toprule
		Method            & train MAE     & val MAE       \\ \hline
		Graph Transformer~\cite{dwivedi2021generalization}   & 0.0944 & 0.1400  \\ 
		Graph Transformer-{\scriptsize Wide}~\cite{dwivedi2021generalization}   & 0.0955 & 0.1408 \\ 
		Graphormer$_{\small \textsc{Small}}$~\cite{ying2021do}  &  0.0778  & \textbf{0.1264}   \\ 
		Graphormer~\cite{ying2021do}   & \textbf{0.0582} & \textbf{0.1234}    \\
		
		$\text{Meta-Transformer-B16}_\text{F}~\cite{zhang2023meta}$   & 0.8034 & 0.8863  \\ 
  \hline
  $\text{Ours}$  & \uline{0.0594} & \uline{0.1397}  \\ 
		\bottomrule
	\end{tabular}
 }
 \caption{\textbf{Graph data understanding }. MAE on PCQM4M-LSC dataset.}
 \label{tab:graph_data}

 }
\end{table*}

\vspace{-0.5em}
\subsection{Adaptation on unseen datasets}
In Table~\ref{tab:generalization} (rows 1-6), we observe that our method performs close to SoTA on all the datasets. Specifically, except on UCF-101, we outperform the SoTA (OmniVec) on all the datasets.  We observe that on NYUv2, we obtain a performance improvement of $3\%$, while on an average perform better by approx $1\%$ on other datasets. It must be noted that we freeze the base embeddings, and unlike other methods do not fine tune the full network, and use simpler task head for analysis on these datasets.  

\subsection{Cross domain generalization}
Table \ref{tab:generalization} (rows 7,8) demonstrates results using our pretrained network on various tasks. On the YouCook2 dataset, our pretrained network surpasses the state of the art in zero-shot retrieval, achieving a Recall@10 of 69.9\% compared to OmniVec's 64.2\% on pretrained network. Interestingly, we are very close to the full fine tuned OmniVec \ie 70.8. This demonstrates that our method is able to leverage the cross domain information better potentially due to multi task  pretraining while OmniVec sequentially trains on one modality at a time. On MSR-VTT, when compared with SM~\cite{zeng2022socratic}, our fine-tuned method has a Recall@10 of 89.4\% cf. SM's 80.0\% (pretrained). It must be noted that SM uses internet-scale data while our method utilizes significantly less data.

\subsection{Adaptation on Unseen Modalities}
\vspace{0.5em} 

\noindent\textbf{Infrared, Hyperspectral, and X-Ray data.}	
Table~\ref{tab:infrared} presents the performance comparison on the RegDB dataset~\cite{nguyen2017person} for infrared image recognition. Our method achieves state of the art performance \ie R@1 of 86.21 c.f. 83.86 of MSCLNet, and mAP of 84.24 c.f. 78.57 of SMCL. This demonstrates that our method can transfer knowledge across unseen modalities. Specifically, we significantly outperform Meta-Transformer, which pretrains on similar modalities as ours. This could be potentially due to separate tokenizers for each modality allowing better integration with the transformer encoder as compared to a common tokenizer in meta-transformer. 

In addition, Table~\ref{tab:hyper} presents the performance on the Indian Pine dataset for hyperspectral image recognition. We achieve an overall accuracy of 90.6\%, which is better than the SpectralFormer (81.76\%) and significantly better than Meta-Transformer(67.6\%).
For X-Ray images (table in supplementary), our method achieves an accuracy of 98.1\%, significantly outperforming competing methods. 
\vspace{0.5em} \\
\noindent\textbf{Graph and IMU Data.}
We show results in Table \ref{tab:graph_data}. We achieve performance close to the state of the art methods \ie validate MAE of 0.1397 c.f. 0.1234 of Graphormer. It is important to note that our method was not designed for graphical data, while competing methods are designed to exploit graphical data. Meta-Transformer, which is a unified learning mechanism like ours, significantly lies behind with 0.8863 MAE cf. 0.1397 of ours. 
\vspace{0.5em} \\
\noindent\textbf{Time series forecasting.} We achieve an MSE of 0.399, 0.601, 0.210, 0.330 on ETTh1, Traffic, Weather and Exchange datasets respectively, outperforming all the competing methods such as Pyraformer~\cite{liu2021pyraformer}, Informer~\cite{haoyietal-informer-2021}, LogTrans~\cite{2019Enhancing}, Meta-former~\cite{yu2022metaformer} and Reformer~\cite{kitaev2020reformer}. The detailed results are in supplementary.

\vspace{0.5em} 

\noindent\textbf{Tabular Data.} We achieve an accuracy of 88.1 and 92.3 on Adult and Bank Marketing datasets respectively, outperforming the competing methods (details in supplementary). Our method has never seen tabular data or structured textual information demonstrating its generalization ability to adapt to unseen patterns within data while providing better performance than competing methods.

\subsection{Ablations}\label{subsec:ablations}
We study the impact of various components of the network in Table \ref{tab:ablations} on image (iNaturalist), video (Kinetics-400) and audio (ESC50) modalities. Specifically, we study the impact of pretraining with a single modality only, using the full pretraining mechanism, and then fine tuning on the respective training set. We also study the impact of modality specific tokenizers compared to unified tokenizers of MetaFormer~\cite{yu2022metaformer}, and impact of utilizing multiple task heads as compared to unified task head design of UniPerceiver-v2~\cite{li2023uni}. For unimodal pretraining (Table \ref{tab:ablations}-row 1), we train the network on a single modality following Step 1 of Masked pretraining (see Sec.~\ref{subsec:training}). We use corresponding modality for each dataset \ie  for iNaturalist, we pretrain on ImageNet1K, for K400, we pretrain on SSv2 and for ESC50, we pretrain on AudioSet. For multimodal multitask pretraining (Table \ref{tab:ablations}-row 2), we pretrain using the full pretraining discussed previously. For fine tuning, we utilize the respective train sets.
\vspace{0.5em} \\
\noindent\textbf{Impact of unimodal vs. multimodal pretraining} We can observe that multimodal multitask pretraining using our approach (row 5) provides a significant improvement in comparison to unimodal pretraining (row 1). Specifically, it outperforms unimodal pretraining by $\sim16\%$ on iNaturalist and K400 datasets while is better by $\sim8\%$ on ESC50. This demonstrates that the network is able to leverage the information from multiple modalities.
\vspace{0.5em} \\
\noindent\textbf{Impact of modality specific tokenizer vs. unified tokenizer.} We observe that the performance of unified tokenizer (row 3) lags behind that of a modality specific tokenizer (row 4) by an average of $\sim5\%$ across all the tasks, while keeping unified heads. Similarly, while keeping task specific heads, and modality specific tokenizer (row 6) vs unified tokenizer (row 5), we observe an average performance gap of $4\%$ in favour of modality specific tokenizer. 
\vspace{0.5em} \\
\noindent\textbf{Multiple task heads vs unified task head.} Comparing row 4 and row 6, we see that the the task specific heads contribute to an increase (average $3.5\%$) in performance while keeping a modality specific tokenizer. 

\begin{table}[]
\resizebox{\columnwidth}{!}{
\begin{tabular}{lllccc}
\toprule
\textbf{Modality} & \textbf{Tokenizer} & \textbf{Task Head} & \textbf{iN2018} & \textbf{K400} & \textbf{ESC50} \\ \midrule
Single            & Modality                               & Autoencoder                            & 74.2            & 78.6          & 82.4           \\
Single            & Unified~\cite{zhang2023meta}        & Autoencoder                            & 74.1            & 78.3          & 82.1           \\
Multiple          & Unified~\cite{zhang2023meta}        & Unified~\cite{li2023uni}               & 80.1            & 81.8          & 82.7           \\
Multiple          & Modality                               & Unified~\cite{li2023uni}               & 85.4            & 84.8          & 86.8           \\
Multiple          & Unified~\cite{zhang2023meta}        & Task specific                          & 86.1            & 85.2          & 87.0           \\
Multiple          & Modality                               & Task specific                          & 90.3            & 88.4          & 92.4           \\ \bottomrule
\end{tabular}
}
\caption{\textbf{Ablation experiments.} We vary the different components of the network to study the impact (Sec~\ref{subsec:ablations}). Metric reported is top-1 accuracy.}
\label{tab:ablations}
\end{table}
\vspace{-0.6em}
\section{Conclusion}
We presented a novel multimodal multitask network and associated training algorithm. The proposed method utilizes modality specific tokenizers and then uses shared transformers based backbone feeding to task specific heads. The traning proceeds in three stages, (i) masked pretraining with one modality at a time, (ii) masked pretraining with pairs of modalities together, and (iii) supervised traning for tasks with pairs of modalities together. The pairwise pretraining and supervised training allows for knowledge sharing between tasks and modalities and leads to a robust and regularized network. We showed empirical results on 25 challenging benchmark datasets over 12 modalities obtaining better or close to existing state of the art results. The method can incorporate arbitrary number of modalities, with only the tokenizer and task heads being modality specific.

{
    \small
    \bibliographystyle{ieeenat_fullname}
    \bibliography{main}

\begin{thebibliography}{105}
\providecommand{\natexlab}[1]{#1}
\providecommand{\url}[1]{\texttt{#1}}
\expandafter\ifx\csname urlstyle\endcsname\relax
  \providecommand{\doi}[1]{doi: #1}\else
  \providecommand{\doi}{doi: \begingroup \urlstyle{rm}\Url}\fi

\bibitem[Akbari et~al.(2021)Akbari, Yuan, Qian, Chuang, Chang, Cui, and Gong]{akbari2021vatt}
Hassan Akbari, Liangzhe Yuan, Rui Qian, Wei-Hong Chuang, Shih-Fu Chang, Yin Cui, and Boqing Gong.
\newblock Vatt: Transformers for multimodal self-supervised learning from raw video, audio and text.
\newblock \emph{Advances in Neural Information Processing Systems}, 34:\penalty0 24206--24221, 2021.

\bibitem[Arandjelovic and Zisserman(2018)]{arandjelovic2018objects}
Relja Arandjelovic and Andrew Zisserman.
\newblock Objects that sound.
\newblock In \emph{Proceedings of the European conference on computer vision (ECCV)}, pages 435--451, 2018.

\bibitem[Armeni et~al.(2016)Armeni, Sener, Zamir, Jiang, Brilakis, Fischer, and Savarese]{armeni20163d}
Iro Armeni, Ozan Sener, Amir~R Zamir, Helen Jiang, Ioannis Brilakis, Martin Fischer, and Silvio Savarese.
\newblock 3d semantic parsing of large-scale indoor spaces.
\newblock In \emph{Proceedings of the IEEE conference on computer vision and pattern recognition}, pages 1534--1543, 2016.

\bibitem[Baade et~al.(2022)Baade, Peng, and Harwath]{baade2022mae}
Alan Baade, Puyuan Peng, and David Harwath.
\newblock Mae-ast: Masked autoencoding audio spectrogram transformer.
\newblock \emph{arXiv preprint arXiv:2203.16691}, 2022.

\bibitem[Baevski et~al.(2022)Baevski, Hsu, Xu, Babu, Gu, and Auli]{baevski2022data2vec}
Alexei Baevski, Wei-Ning Hsu, Qiantong Xu, Arun Babu, Jiatao Gu, and Michael Auli.
\newblock Data2vec: A general framework for self-supervised learning in speech, vision and language.
\newblock In \emph{International Conference on Machine Learning}, pages 1298--1312. PMLR, 2022.

\bibitem[Bai et~al.(2022)Bai, Lu, Xiao, Chen, and Jiao]{bai2022generative}
Jing Bai, Jiawei Lu, Zhu Xiao, Zheng Chen, and Licheng Jiao.
\newblock Generative adversarial networks based on transformer encoder and convolution block for hyperspectral image classification.
\newblock \emph{Remote Sensing}, 14\penalty0 (14):\penalty0 3426, 2022.

\bibitem[Carreira et~al.(2022)Carreira, Koppula, Zoran, Recasens, Ionescu, Henaff, Shelhamer, Arandjelovic, Botvinick, Vinyals, et~al.]{carreira2022hierarchical}
Joao Carreira, Skanda Koppula, Daniel Zoran, Adria Recasens, Catalin Ionescu, Olivier Henaff, Evan Shelhamer, Relja Arandjelovic, Matt Botvinick, Oriol Vinyals, et~al.
\newblock Hierarchical perceiver.
\newblock \emph{arXiv preprint arXiv:2202.10890}, 2022.

\bibitem[Chen et~al.(2022)Chen, Du, Zhu, Ma, Berg-Kirkpatrick, and Dubnov]{chen2022hts}
Ke Chen, Xingjian Du, Bilei Zhu, Zejun Ma, Taylor Berg-Kirkpatrick, and Shlomo Dubnov.
\newblock Hts-at: A hierarchical token-semantic audio transformer for sound classification and detection.
\newblock In \emph{ICASSP 2022-2022 IEEE International Conference on Acoustics, Speech and Signal Processing (ICASSP)}, pages 646--650. IEEE, 2022.

\bibitem[Chen et~al.(2021)Chen, Liu, Chen, and Zhang]{chen2021dialogsum}
Yulong Chen, Yang Liu, Liang Chen, and Yue Zhang.
\newblock Dialogsum: A real-life scenario dialogue summarization dataset.
\newblock \emph{arXiv preprint arXiv:2105.06762}, 2021.

\bibitem[Crawshaw(2020)]{mtlsurvey2020}
Michael Crawshaw.
\newblock Multi-task learning with deep neural networks: {A} survey.
\newblock \emph{arXiv preprint arXiv:2009.09796}, 2020.

\bibitem[Dai et~al.(2022)Dai, Tang, Liu, Tan, Zhou, Wang, Feng, Zhang, Hu, and Shi]{dai2022one}
Yong Dai, Duyu Tang, Liangxin Liu, Minghuan Tan, Cong Zhou, Jingquan Wang, Zhangyin Feng, Fan Zhang, Xueyu Hu, and Shuming Shi.
\newblock One model, multiple modalities: A sparsely activated approach for text, sound, image, video and code.
\newblock \emph{arXiv preprint arXiv:2205.06126}, 2022.

\bibitem[Deng et~al.(2009)Deng, Dong, Socher, Li, Li, and Fei-Fei]{deng2009imagenet}
Jia Deng, Wei Dong, Richard Socher, Li-Jia Li, Kai Li, and Li Fei-Fei.
\newblock Imagenet: A large-scale hierarchical image database.
\newblock In \emph{CVPR}, pages 248--255. Ieee, 2009.

\bibitem[Devlin et~al.(2019)Devlin, Chang, Lee, and Toutanova]{devlin2018bert}
Jacob Devlin, Ming-Wei Chang, Kenton Lee, and Kristina Toutanova.
\newblock {BERT}: Pre-training of deep bidirectional transformers for language understanding.
\newblock In \emph{NAACL-HLT}, 2019.

\bibitem[Dosovitskiy et~al.(2020)Dosovitskiy, Beyer, Kolesnikov, Weissenborn, Zhai, Unterthiner, Dehghani, Minderer, Heigold, Gelly, et~al.]{dosovitskiy2020image}
Alexey Dosovitskiy, Lucas Beyer, Alexander Kolesnikov, Dirk Weissenborn, Xiaohua Zhai, Thomas Unterthiner, Mostafa Dehghani, Matthias Minderer, Georg Heigold, Sylvain Gelly, et~al.
\newblock An image is worth 16x16 words: Transformers for image recognition at scale.
\newblock \emph{arXiv preprint arXiv:2010.11929}, 2020.

\bibitem[Dwivedi and Bresson(2021)]{dwivedi2021generalization}
Vijay~Prakash Dwivedi and Xavier Bresson.
\newblock A generalization of transformer networks to graphs.
\newblock \emph{AAAI Workshop on Deep Learning on Graphs: Methods and Applications}, 2021.

\bibitem[Fifty et~al.(2021)Fifty, Amid, Zhao, Yu, Anil, and Finn]{fifty2021efficiently}
Chris Fifty, Ehsan Amid, Zhe Zhao, Tianhe Yu, Rohan Anil, and Chelsea Finn.
\newblock Efficiently identifying task groupings for multi-task learning.
\newblock \emph{Advances in Neural Information Processing Systems}, 34:\penalty0 27503--27516, 2021.

\bibitem[Gazneli et~al.(2022)Gazneli, Zimerman, Ridnik, Sharir, and Noy]{gazneli2022end}
Avi Gazneli, Gadi Zimerman, Tal Ridnik, Gilad Sharir, and Asaf Noy.
\newblock End-to-end audio strikes back: Boosting augmentations towards an efficient audio classification network.
\newblock \emph{arXiv preprint arXiv:2204.11479}, 2022.

\bibitem[Gemmeke et~al.(2017)Gemmeke, Ellis, Freedman, Jansen, Lawrence, Moore, Plakal, and Ritter]{gemmeke2017audio}
Jort~F Gemmeke, Daniel~PW Ellis, Dylan Freedman, Aren Jansen, Wade Lawrence, R~Channing Moore, Manoj Plakal, and Marvin Ritter.
\newblock Audio set: An ontology and human-labeled dataset for audio events.
\newblock In \emph{2017 IEEE international conference on acoustics, speech and signal processing (ICASSP)}, pages 776--780. IEEE, 2017.

\bibitem[Girdhar et~al.(2022{\natexlab{a}})Girdhar, El-Nouby, Singh, Alwala, Joulin, and Misra]{girdhar2022omnimae}
Rohit Girdhar, Alaaeldin El-Nouby, Mannat Singh, Kalyan~Vasudev Alwala, Armand Joulin, and Ishan Misra.
\newblock Omnimae: Single model masked pretraining on images and videos.
\newblock \emph{arXiv preprint arXiv:2206.08356}, 2022{\natexlab{a}}.

\bibitem[Girdhar et~al.(2022{\natexlab{b}})Girdhar, Singh, Ravi, van~der Maaten, Joulin, and Misra]{girdhar2022omnivore}
Rohit Girdhar, Mannat Singh, Nikhila Ravi, Laurens van~der Maaten, Armand Joulin, and Ishan Misra.
\newblock Omnivore: A single model for many visual modalities.
\newblock In \emph{Proceedings of the IEEE/CVF Conference on Computer Vision and Pattern Recognition}, pages 16102--16112, 2022{\natexlab{b}}.

\bibitem[Gliwa et~al.(2019)Gliwa, Mochol, Biesek, and Wawer]{gliwa2019samsum}
Bogdan Gliwa, Iwona Mochol, Maciej Biesek, and Aleksander Wawer.
\newblock Samsum corpus: A human-annotated dialogue dataset for abstractive summarization.
\newblock \emph{arXiv preprint arXiv:1911.12237}, 2019.

\bibitem[Gong et~al.(2021)Gong, Chung, and Glass]{gong2021ast}
Yuan Gong, Yu-An Chung, and James Glass.
\newblock Ast: Audio spectrogram transformer.
\newblock \emph{arXiv preprint arXiv:2104.01778}, 2021.

\bibitem[Gong et~al.(2022)Gong, Liu, Rouditchenko, and Glass]{gong2022uavm}
Yuan Gong, Alexander~H Liu, Andrew Rouditchenko, and James Glass.
\newblock Uavm: Towards unifying audio and visual models.
\newblock \emph{IEEE Signal Processing Letters}, 29:\penalty0 2437--2441, 2022.

\bibitem[Goyal et~al.(2017)Goyal, Ebrahimi~Kahou, Michalski, Materzynska, Westphal, Kim, Haenel, Fruend, Yianilos, Mueller-Freitag, et~al.]{goyal2017something}
Raghav Goyal, Samira Ebrahimi~Kahou, Vincent Michalski, Joanna Materzynska, Susanne Westphal, Heuna Kim, Valentin Haenel, Ingo Fruend, Peter Yianilos, Moritz Mueller-Freitag, et~al.
\newblock The" something something" video database for learning and evaluating visual common sense.
\newblock In \emph{Proceedings of the IEEE international conference on computer vision}, pages 5842--5850, 2017.

\bibitem[Grauman et~al.(2022)Grauman, Westbury, Byrne, Chavis, Furnari, Girdhar, Hamburger, Jiang, Liu, Liu, et~al.]{grauman2022ego4d}
Kristen Grauman, Andrew Westbury, Eugene Byrne, Zachary Chavis, Antonino Furnari, Rohit Girdhar, Jackson Hamburger, Hao Jiang, Miao Liu, Xingyu Liu, et~al.
\newblock Ego4d: Around the world in 3,000 hours of egocentric video.
\newblock In \emph{Proceedings of the IEEE/CVF Conference on Computer Vision and Pattern Recognition}, pages 18995--19012, 2022.

\bibitem[Guo et~al.(2018)Guo, Haque, Huang, Yeung, and Fei-Fei]{guo2018dynamic}
Michelle Guo, Albert Haque, De-An Huang, Serena Yeung, and Li Fei-Fei.
\newblock Dynamic task prioritization for multitask learning.
\newblock In \emph{Proceedings of the European conference on computer vision (ECCV)}, pages 270--287, 2018.

\bibitem[Gupta et~al.(2022)Gupta, Tian, Zhang, Wu, Martin, and Fei-Fei]{gupta2022maskvit}
Agrim Gupta, Stephen Tian, Yunzhi Zhang, Jiajun Wu, Roberto Martin, and Li Fei-Fei.
\newblock Maskvit: Masked visual pre-training for video prediction.
\newblock \emph{arXiv preprint arXiv:2206.11894}, 2022.

\bibitem[He et~al.(2022)He, Chen, Xie, Li, Dollar, and Girshick]{he2022masked}
Kaiming He, Xinlei Chen, Saining Xie, Yanghao Li, Piotr Dollar, and Ross Girshick.
\newblock Masked autoencoders are scalable vision learners.
\newblock In \emph{Proceedings of the IEEE/CVF Conference on Computer Vision and Pattern Recognition}, pages 16000--16009, 2022.

\bibitem[Hendrycks and Gimpel(2016)]{hendrycks2016gaussian}
Dan Hendrycks and Kevin Gimpel.
\newblock Gaussian error linear units (gelus).
\newblock \emph{arXiv preprint arXiv:1606.08415}, 2016.

\bibitem[Hong et~al.(2021)Hong, Han, Yao, Gao, Zhang, Plaza, and Chanussot]{hong2021spectralformer}
Danfeng Hong, Zhu Han, Jing Yao, Lianru Gao, Bing Zhang, Antonio Plaza, and Jocelyn Chanussot.
\newblock Spectralformer: Rethinking hyperspectral image classification with transformers.
\newblock \emph{IEEE Transactions on Geoscience and Remote Sensing}, 60:\penalty0 1--15, 2021.

\bibitem[Hu and Singh(2021)]{hu2021unit}
Ronghang Hu and Amanpreet Singh.
\newblock Unit: Multimodal multitask learning with a unified transformer.
\newblock In \emph{Proceedings of the IEEE/CVF International Conference on Computer Vision}, pages 1439--1449, 2021.

\bibitem[Hu et~al.(2021)Hu, Fey, Ren, Nakata, Dong, and Leskovec]{hu2021ogb}
Weihua Hu, Matthias Fey, Hongyu Ren, Maho Nakata, Yuxiao Dong, and Jure Leskovec.
\newblock Ogb-lsc: A large-scale challenge for machine learning on graphs.
\newblock \emph{arXiv preprint arXiv:2103.09430}, 2021.

\bibitem[Jaegle et~al.(2021{\natexlab{a}})Jaegle, Borgeaud, Alayrac, Doersch, Ionescu, Ding, Koppula, Zoran, Brock, Shelhamer, et~al.]{jaegle2021perceiverio}
Andrew Jaegle, Sebastian Borgeaud, Jean-Baptiste Alayrac, Carl Doersch, Catalin Ionescu, David Ding, Skanda Koppula, Daniel Zoran, Andrew Brock, Evan Shelhamer, et~al.
\newblock Perceiver io: A general architecture for structured inputs \& outputs.
\newblock \emph{arXiv preprint arXiv:2107.14795}, 2021{\natexlab{a}}.

\bibitem[Jaegle et~al.(2021{\natexlab{b}})Jaegle, Gimeno, Brock, Vinyals, Zisserman, and Carreira]{jaegle2021perceiver}
Andrew Jaegle, Felix Gimeno, Andy Brock, Oriol Vinyals, Andrew Zisserman, and Joao Carreira.
\newblock Perceiver: General perception with iterative attention.
\newblock In \emph{International conference on machine learning}, pages 4651--4664. PMLR, 2021{\natexlab{b}}.

\bibitem[Jiang et~al.(2021)Jiang, Ma, Xiao, Shao, and Guo]{jiang2021review}
Xingyu Jiang, Jiayi Ma, Guobao Xiao, Zhenfeng Shao, and Xiaojie Guo.
\newblock A review of multimodal image matching: Methods and applications.
\newblock \emph{Information Fusion}, 73:\penalty0 22--71, 2021.

\bibitem[Kaiser et~al.(2017)Kaiser, Gomez, Shazeer, Vaswani, Parmar, Jones, and Uszkoreit]{kaiser2017one}
Lukasz Kaiser, Aidan~N Gomez, Noam Shazeer, Ashish Vaswani, Niki Parmar, Llion Jones, and Jakob Uszkoreit.
\newblock One model to learn them all.
\newblock \emph{arXiv preprint arXiv:1706.05137}, 2017.

\bibitem[Kay et~al.(2017)Kay, Carreira, Simonyan, Zhang, Hillier, Vijayanarasimhan, Viola, Green, Back, Natsev, et~al.]{kay2017kinetics}
Will Kay, Joao Carreira, Karen Simonyan, Brian Zhang, Chloe Hillier, Sudheendra Vijayanarasimhan, Fabio Viola, Tim Green, Trevor Back, Paul Natsev, et~al.
\newblock The kinetics human action video dataset.
\newblock \emph{arXiv preprint arXiv:1705.06950}, 2017.

\bibitem[Kim et~al.(2022)Kim, Joo, Chae, Kim, Hwang, and Yeo]{kim2022mind}
Seungone Kim, Se~June Joo, Hyungjoo Chae, Chaehyeong Kim, Seung-won Hwang, and Jinyoung Yeo.
\newblock Mind the gap! injecting commonsense knowledge for abstractive dialogue summarization.
\newblock \emph{arXiv preprint arXiv:2209.00930}, 2022.

\bibitem[Kitaev et~al.(2020)Kitaev, Kaiser, and Levskaya]{kitaev2020reformer}
Nikita Kitaev, Lukasz Kaiser, and Anselm Levskaya.
\newblock Reformer: The efficient transformer.
\newblock In \emph{ICLR}, 2020.

\bibitem[Kuehne et~al.(2011)Kuehne, Jhuang, Garrote, Poggio, and Serre]{kuehne2011hmdb}
Hildegard Kuehne, Hueihan Jhuang, Estibaliz Garrote, Tomaso Poggio, and Thomas Serre.
\newblock Hmdb: a large video database for human motion recognition.
\newblock In \emph{2011 International conference on computer vision}, pages 2556--2563. IEEE, 2011.

\bibitem[Lai et~al.(2018)Lai, Chang, Yang, and Liu]{lai2018modeling}
Guokun Lai, Wei-Cheng Chang, Yiming Yang, and Hanxiao Liu.
\newblock Modeling long-and short-term temporal patterns with deep neural networks.
\newblock In \emph{The 41st international ACM SIGIR conference on research \& development in information retrieval}, pages 95--104, 2018.

\bibitem[Lai et~al.(2022)Lai, Liu, Jiang, Wang, Zhao, Liu, Qi, and Jia]{lai2022stratified}
Xin Lai, Jianhui Liu, Li Jiang, Liwei Wang, Hengshuang Zhao, Shu Liu, Xiaojuan Qi, and Jiaya Jia.
\newblock Stratified transformer for 3d point cloud segmentation.
\newblock In \emph{Proceedings of the IEEE/CVF Conference on Computer Vision and Pattern Recognition}, pages 8500--8509, 2022.

\bibitem[Lee et~al.(2021)Lee, Lee, Lee, Lee, Lee, Woo, and Lee]{lee2021regularization}
Dogyoon Lee, Jaeha Lee, Junhyeop Lee, Hyeongmin Lee, Minhyeok Lee, Sungmin Woo, and Sangyoun Lee.
\newblock Regularization strategy for point cloud via rigidly mixed sample.
\newblock In \emph{Proceedings of the IEEE/CVF Conference on Computer Vision and Pattern Recognition}, pages 15900--15909, 2021.

\bibitem[Li et~al.(2023)Li, Zhu, Jiang, Zhu, Li, Yuan, Wang, Qiao, Wang, Wang, et~al.]{li2023uni}
Hao Li, Jinguo Zhu, Xiaohu Jiang, Xizhou Zhu, Hongsheng Li, Chun Yuan, Xiaohua Wang, Yu Qiao, Xiaogang Wang, Wenhai Wang, et~al.
\newblock Uni-perceiver v2: A generalist model for large-scale vision and vision-language tasks.
\newblock In \emph{Proceedings of the IEEE/CVF Conference on Computer Vision and Pattern Recognition}, pages 2691--2700, 2023.

\bibitem[Li et~al.(2022)Li, Wang, He, Li, Wang, Wang, and Qiao]{li2022uniformerv2}
Kunchang Li, Yali Wang, Yinan He, Yizhuo Li, Yi Wang, Limin Wang, and Yu Qiao.
\newblock Uniformerv2: Spatiotemporal learning by arming image vits with video uniformer.
\newblock \emph{arXiv preprint arXiv:2211.09552}, 2022.

\bibitem[Li et~al.(2019)Li, Jin, Xuan, Zhou, Chen, Wang, and Yan]{2019Enhancing}
Shiyang Li, Xiaoyong Jin, Yao Xuan, Xiyou Zhou, Wenhu Chen, Yu-Xiang Wang, and Xifeng Yan.
\newblock Enhancing the locality and breaking the memory bottleneck of transformer on time series forecasting.
\newblock In \emph{NeurIPS}, 2019.

\bibitem[Liang et~al.(2022)Liang, Zhang, Kwon, Yeung, and Zou]{liang2022mind}
Weixin Liang, Yuhui Zhang, Yongchan Kwon, Serena Yeung, and James Zou.
\newblock Mind the gap: Understanding the modality gap in multi-modal contrastive representation learning.
\newblock \emph{arXiv preprint arXiv:2203.02053}, 2022.

\bibitem[Lin et~al.(2014)Lin, Maire, Belongie, Hays, Perona, Ramanan, Doll{\'{a}}r, and Zitnick]{lin2014microsoft}
Tsung{-}Yi Lin, Michael Maire, Serge~J. Belongie, James Hays, Pietro Perona, Deva Ramanan, Piotr Doll{\'{a}}r, and C.~Lawrence Zitnick.
\newblock Microsoft coco: Common objects in context.
\newblock In \emph{ECCV}, 2014.

\bibitem[Liu et~al.(2021{\natexlab{a}})Liu, Zhu, Liu, Guo, Zhao, Sun, Wang, Lu, Zhou, Zhang, et~al.]{liu2021opt}
Jing Liu, Xinxin Zhu, Fei Liu, Longteng Guo, Zijia Zhao, Mingzhen Sun, Weining Wang, Hanqing Lu, Shiyu Zhou, Jiajun Zhang, et~al.
\newblock Opt: Omni-perception pre-trainer for cross-modal understanding and generation.
\newblock \emph{arXiv preprint arXiv:2107.00249}, 2021{\natexlab{a}}.

\bibitem[Liu et~al.(2021{\natexlab{b}})Liu, Yu, Liao, Li, Lin, Liu, and Dustdar]{liu2021pyraformer}
Shizhan Liu, Hang Yu, Cong Liao, Jianguo Li, Weiyao Lin, Alex~X Liu, and Schahram Dustdar.
\newblock Pyraformer: Low-complexity pyramidal attention for long-range time series modeling and forecasting.
\newblock In \emph{ICLR}, 2021{\natexlab{b}}.

\bibitem[Liu et~al.(2019)Liu, Ott, Goyal, Du, Joshi, Chen, Levy, Lewis, Zettlemoyer, and Stoyanov]{liu2019roberta}
Yinhan Liu, Myle Ott, Naman Goyal, Jingfei Du, Mandar Joshi, Danqi Chen, Omer Levy, Mike Lewis, Luke Zettlemoyer, and Veselin Stoyanov.
\newblock Roberta: A robustly optimized bert pretraining approach.
\newblock \emph{arXiv preprint arXiv:1907.11692}, 2019.

\bibitem[Monfort et~al.(2019)Monfort, Andonian, Zhou, Ramakrishnan, Bargal, Yan, Brown, Fan, Gutfreund, Vondrick, et~al.]{monfort2019moments}
Mathew Monfort, Alex Andonian, Bolei Zhou, Kandan Ramakrishnan, Sarah~Adel Bargal, Tom Yan, Lisa Brown, Quanfu Fan, Dan Gutfreund, Carl Vondrick, et~al.
\newblock Moments in time dataset: one million videos for event understanding.
\newblock \emph{IEEE transactions on pattern analysis and machine intelligence}, 42\penalty0 (2):\penalty0 502--508, 2019.

\bibitem[Nguyen et~al.(2017)Nguyen, Hong, Kim, and Park]{nguyen2017person}
Dat~Tien Nguyen, Hyung~Gil Hong, Ki~Wan Kim, and Kang~Ryoung Park.
\newblock Person recognition system based on a combination of body images from visible light and thermal cameras.
\newblock \emph{Sensors}, 17\penalty0 (3):\penalty0 605, 2017.

\bibitem[Oreshkin et~al.(2019)Oreshkin, Carpov, Chapados, and Bengio]{oreshkin2019n}
Boris~N Oreshkin, Dmitri Carpov, Nicolas Chapados, and Yoshua Bengio.
\newblock N-beats: Neural basis expansion analysis for interpretable time series forecasting.
\newblock \emph{arXiv preprint arXiv:1905.10437}, 2019.

\bibitem[Parkhi et~al.(2012)Parkhi, Vedaldi, Zisserman, and Jawahar]{parkhi2012cats}
Omkar~M Parkhi, Andrea Vedaldi, Andrew Zisserman, and CV Jawahar.
\newblock Cats and dogs.
\newblock In \emph{2012 IEEE conference on computer vision and pattern recognition}, pages 3498--3505. IEEE, 2012.

\bibitem[Piczak(2015)]{piczak2015esc}
Karol~J Piczak.
\newblock Esc: Dataset for environmental sound classification.
\newblock In \emph{Proceedings of the 23rd ACM international conference on Multimedia}, pages 1015--1018, 2015.

\bibitem[Piergiovanni et~al.(2023)Piergiovanni, Kuo, and Angelova]{piergiovanni2023rethinking}
AJ Piergiovanni, Weicheng Kuo, and Anelia Angelova.
\newblock Rethinking video vits: Sparse video tubes for joint image and video learning.
\newblock In \emph{Proceedings of the IEEE/CVF Conference on Computer Vision and Pattern Recognition}, pages 2214--2224, 2023.

\bibitem[Pramanik et~al.(2019)Pramanik, Agrawal, and Hussain]{pramanik2019omninet}
Subhojeet Pramanik, Priyanka Agrawal, and Aman Hussain.
\newblock Omninet: A unified architecture for multi-modal multi-task learning.
\newblock \emph{arXiv preprint arXiv:1907.07804}, 2019.

\bibitem[Qi et~al.(2017)Qi, Yi, Su, and Guibas]{qi2017pointnet++}
Charles~R Qi, Li Yi, Hao Su, and Leonidas~J Guibas.
\newblock Pointnet++: Deep hierarchical feature learning on point sets in a metric space.
\newblock In \emph{NeurIPS}, 2017.

\bibitem[Radford et~al.(2018)Radford, Narasimhan, Salimans, Sutskever, et~al.]{radford2018improving}
Alec Radford, Karthik Narasimhan, Tim Salimans, Ilya Sutskever, et~al.
\newblock Improving language understanding by generative pre-training.
\newblock 2018.

\bibitem[Rahman et~al.(2020)Rahman, Khandakar, Kadir, Islam, Islam, Mazhar, Hamid, Islam, Kashem, Mahbub, et~al.]{rahman2020reliable}
Tawsifur Rahman, Amith Khandakar, Muhammad~Abdul Kadir, Khandaker~Rejaul Islam, Khandakar~F Islam, Rashid Mazhar, Tahir Hamid, Mohammad~Tariqul Islam, Saad Kashem, Zaid~Bin Mahbub, et~al.
\newblock Reliable tuberculosis detection using chest x-ray with deep learning, segmentation and visualization.
\newblock \emph{IEEE Access}, 8:\penalty0 191586--191601, 2020.

\bibitem[Recasens et~al.(2023)Recasens, Lin, Carreira, Jaegle, Wang, Alayrac, Luc, Miech, Smaira, Hemsley, et~al.]{recasens2023zorro}
Adria Recasens, Jason Lin, Joao Carreira, Drew Jaegle, Luyu Wang, Jean-baptiste Alayrac, Pauline Luc, Antoine Miech, Lucas Smaira, Ross Hemsley, et~al.
\newblock Zorro: the masked multimodal transformer.
\newblock \emph{arXiv preprint arXiv:2301.09595}, 2023.

\bibitem[Silberman et~al.(2012)Silberman, Hoiem, Kohli, and Fergus]{silberman2012indoor}
Nathan Silberman, Derek Hoiem, Pushmeet Kohli, and Rob Fergus.
\newblock Indoor segmentation and support inference from rgbd images.
\newblock \emph{ECCV (5)}, 7576:\penalty0 746--760, 2012.

\bibitem[Song et~al.(2020)Song, Tan, Qin, Lu, and Liu]{song2020mpnet}
Kaitao Song, Xu Tan, Tao Qin, Jianfeng Lu, and Tie-Yan Liu.
\newblock Mpnet: Masked and permuted pre-training for language understanding.
\newblock \emph{Advances in Neural Information Processing Systems}, 33:\penalty0 16857--16867, 2020.

\bibitem[Song et~al.(2015)Song, Lichtenberg, and Xiao]{song2015sun}
Shuran Song, Samuel~P Lichtenberg, and Jianxiong Xiao.
\newblock Sun rgb-d: A rgb-d scene understanding benchmark suite.
\newblock In \emph{Proceedings of the IEEE conference on computer vision and pattern recognition}, pages 567--576, 2015.

\bibitem[Soomro et~al.(2012)Soomro, Zamir, and Shah]{soomro2012ucf101}
Khurram Soomro, Amir~Roshan Zamir, and Mubarak Shah.
\newblock Ucf101: A dataset of 101 human actions classes from videos in the wild.
\newblock \emph{arXiv preprint arXiv:1212.0402}, 2012.

\bibitem[Srivastava and Sharma(2023)]{srivastava2023omnivec}
Siddharth Srivastava and Gaurav Sharma.
\newblock Omnivec: Learning robust representations with cross modal sharing.
\newblock \emph{arXiv preprint arXiv:2311.05709}, 2023.

\bibitem[Srivastava et~al.(2023)Srivastava, Bhugra, Kaushik, and Lall]{srivastava2023hierarchical}
Siddharth Srivastava, Swati Bhugra, Vinay Kaushik, and Brejesh Lall.
\newblock Hierarchical multi-task learning via task affinity groupings.
\newblock In \emph{2023 IEEE International Conference on Image Processing (ICIP)}, pages 3289--3293. IEEE, 2023.

\bibitem[Sun et~al.(2022)Sun, Zhang, Kailkhura, Yu, Xiao, and Mao]{sun2022benchmarking}
Jiachen Sun, Qingzhao Zhang, Bhavya Kailkhura, Zhiding Yu, Chaowei Xiao, and Z~Morley Mao.
\newblock Benchmarking robustness of 3d point cloud recognition against common corruptions.
\newblock \emph{arXiv preprint arXiv:2201.12296}, 2022.

\bibitem[Tan and Le(2019)]{tan2019efficientnet}
Mingxing Tan and Quoc Le.
\newblock Efficientnet: Rethinking model scaling for convolutional neural networks.
\newblock In \emph{International conference on machine learning}, pages 6105--6114. PMLR, 2019.

\bibitem[Tang et~al.(2022)Tang, Zhan, Chen, Yu, and Tao]{tang2022contrastive}
Liyao Tang, Yibing Zhan, Zhe Chen, Baosheng Yu, and Dacheng Tao.
\newblock Contrastive boundary learning for point cloud segmentation.
\newblock In \emph{Proceedings of the IEEE/CVF Conference on Computer Vision and Pattern Recognition}, pages 8489--8499, 2022.

\bibitem[Uchaev and Uchaev(2023)]{uchaev2023small}
Denis Uchaev and Dmitry Uchaev.
\newblock Small sample hyperspectral image classification based on the random patches network and recursive filtering.
\newblock \emph{Sensors}, 23\penalty0 (5):\penalty0 2499, 2023.

\bibitem[Uy et~al.(2019)Uy, Pham, Hua, Nguyen, and Yeung]{uy2019revisiting}
Mikaela~Angelina Uy, Quang-Hieu Pham, Binh-Son Hua, Thanh Nguyen, and Sai-Kit Yeung.
\newblock Revisiting point cloud classification: A new benchmark dataset and classification model on real-world data.
\newblock In \emph{ICCV}, pages 1588--1597, 2019.

\bibitem[Van~Horn et~al.(2018)Van~Horn, Mac~Aodha, Song, Cui, Sun, Shepard, Adam, Perona, and Belongie]{van2018inaturalist}
Grant Van~Horn, Oisin Mac~Aodha, Yang Song, Yin Cui, Chen Sun, Alex Shepard, Hartwig Adam, Pietro Perona, and Serge Belongie.
\newblock The inaturalist species classification and detection dataset.
\newblock In \emph{Proceedings of the IEEE conference on computer vision and pattern recognition}, pages 8769--8778, 2018.

\bibitem[Vaswani et~al.(2017)Vaswani, Shazeer, Parmar, Uszkoreit, Jones, Gomez, Kaiser, and Polosukhin]{vaswani2017attention}
Ashish Vaswani, Noam Shazeer, Niki Parmar, Jakob Uszkoreit, Llion Jones, Aidan~N Gomez, Lukasz Kaiser, and Illia Polosukhin.
\newblock Attention is all you need.
\newblock \emph{Advances in neural information processing systems}, 30, 2017.

\bibitem[Wang et~al.(2023)Wang, Dai, Chen, Huang, Li, Zhu, Hu, Lu, Lu, Li, et~al.]{wang2023internimage}
Wenhai Wang, Jifeng Dai, Zhe Chen, Zhenhang Huang, Zhiqi Li, Xizhou Zhu, Xiaowei Hu, Tong Lu, Lewei Lu, Hongsheng Li, et~al.
\newblock Internimage: Exploring large-scale vision foundation models with deformable convolutions.
\newblock In \emph{Proceedings of the IEEE/CVF Conference on Computer Vision and Pattern Recognition}, pages 14408--14419, 2023.

\bibitem[Wang et~al.(2022)Wang, Li, Li, He, Huang, Zhao, Zhang, Xu, Liu, Wang, et~al.]{wang2022internvideo}
Yi Wang, Kunchang Li, Yizhuo Li, Yinan He, Bingkun Huang, Zhiyu Zhao, Hongjie Zhang, Jilan Xu, Yi Liu, Zun Wang, et~al.
\newblock Internvideo: General video foundation models via generative and discriminative learning.
\newblock \emph{arXiv preprint arXiv:2212.03191}, 2022.

\bibitem[Wei et~al.(2022)Wei, Fan, Xie, Wu, Yuille, and Feichtenhofer]{wei2022masked}
Chen Wei, Haoqi Fan, Saining Xie, Chao-Yuan Wu, Alan Yuille, and Christoph Feichtenhofer.
\newblock Masked feature prediction for self-supervised visual pre-training.
\newblock In \emph{Proceedings of the IEEE/CVF Conference on Computer Vision and Pattern Recognition}, pages 14668--14678, 2022.

\bibitem[Wei et~al.(2021)Wei, Yang, Wang, and Gao]{Wei_2021_ICCV}
Ziyu Wei, Xi Yang, Nannan Wang, and Xinbo Gao.
\newblock Syncretic modality collaborative learning for visible infrared person re-identification.
\newblock In \emph{ICCV}, pages 225--234, 2021.

\bibitem[Wu et~al.(2021)Wu, Liu, Liu, Stenetorp, and Xiong]{wu2021controllable}
Chien-Sheng Wu, Linqing Liu, Wenhao Liu, Pontus Stenetorp, and Caiming Xiong.
\newblock Controllable abstractive dialogue summarization with sketch supervision.
\newblock \emph{arXiv preprint arXiv:2105.14064}, 2021.

\bibitem[Wu et~al.(2022{\natexlab{a}})Wu, Zhang, Peng, Liu, Xiao, Fu, and Yuan]{wu2022tinyvit}
Kan Wu, Jinnian Zhang, Houwen Peng, Mengchen Liu, Bin Xiao, Jianlong Fu, and Lu Yuan.
\newblock Tinyvit: Fast pretraining distillation for small vision transformers.
\newblock In \emph{European Conference on Computer Vision}, pages 68--85. Springer, 2022{\natexlab{a}}.

\bibitem[Wu et~al.(2022{\natexlab{b}})Wu, Lao, Jiang, Liu, and Zhao]{wu2022point}
Xiaoyang Wu, Yixing Lao, Li Jiang, Xihui Liu, and Hengshuang Zhao.
\newblock Point transformer v2: Grouped vector attention and partition-based pooling.
\newblock \emph{Advances in Neural Information Processing Systems}, 35:\penalty0 33330--33342, 2022{\natexlab{b}}.

\bibitem[Wu et~al.(2015)Wu, Song, Khosla, Yu, Zhang, Tang, and Xiao]{wu20153d}
Zhirong Wu, Shuran Song, Aditya Khosla, Fisher Yu, Linguang Zhang, Xiaoou Tang, and Jianxiong Xiao.
\newblock 3d shapenets: A deep representation for volumetric shapes.
\newblock In \emph{Proceedings of the IEEE conference on computer vision and pattern recognition}, pages 1912--1920, 2015.

\bibitem[Xiao et~al.(2020)Xiao, Lee, Grauman, Malik, and Feichtenhofer]{xiao2020audiovisual}
Fanyi Xiao, Yong~Jae Lee, Kristen Grauman, Jitendra Malik, and Christoph Feichtenhofer.
\newblock Audiovisual slowfast networks for video recognition.
\newblock \emph{arXiv preprint arXiv:2001.08740}, 2020.

\bibitem[Xu et~al.(2016)Xu, Mei, Yao, and Rui]{xu2016msr}
Jun Xu, Tao Mei, Ting Yao, and Yong Rui.
\newblock Msr-vtt: A large video description dataset for bridging video and language.
\newblock In \emph{Proceedings of the IEEE conference on computer vision and pattern recognition}, pages 5288--5296, 2016.

\bibitem[Xu et~al.(2022)Xu, Zhu, and Clifton]{xu2022multimodal}
Peng Xu, Xiatian Zhu, and David~A Clifton.
\newblock Multimodal learning with transformers: A survey.
\newblock \emph{arXiv preprint arXiv:2206.06488}, 2022.

\bibitem[Yan et~al.(2022)Yan, Li, Wang, Zhang, Li, and Yang]{yan2022multi}
Zhiqiang Yan, Xiang Li, Kun Wang, Zhenyu Zhang, Jun Li, and Jian Yang.
\newblock Multi-modal masked pre-training for monocular panoramic depth completion.
\newblock In \emph{Computer Vision--ECCV 2022: 17th European Conference, Tel Aviv, Israel, October 23--27, 2022, Proceedings, Part I}, pages 378--395. Springer, 2022.

\bibitem[Yang et~al.(2023)Yang, Guo, Xiong, Liu, Pan, Wang, Tong, and Guo]{yang2023swin3d}
Yu-Qi Yang, Yu-Xiao Guo, Jian-Yu Xiong, Yang Liu, Hao Pan, Peng-Shuai Wang, Xin Tong, and Baining Guo.
\newblock Swin3d: A pretrained transformer backbone for 3d indoor scene understanding.
\newblock \emph{arXiv preprint arXiv:2304.06906}, 2023.

\bibitem[Yang et~al.(2021)Yang, Dai, Yang, Carbonell, Salakhutdinov, and Le~QV]{yang2021generalized}
Z Yang, Z Dai, Y Yang, J Carbonell, RR Salakhutdinov, and XLNet Le~QV.
\newblock generalized autoregressive pretraining for language understanding; 2019.
\newblock \emph{Preprint at https://arxiv. org/abs/1906.08237 Accessed June}, 21, 2021.

\bibitem[Ye et~al.(2020)Ye, Shen, Lin, Xiang, Shao, and Hoi]{arxiv20reidsurvey}
Mang Ye, Jianbing Shen, Gaojie Lin, Tao Xiang, Ling Shao, and Steven C.~H. Hoi.
\newblock Deep learning for person re-identification: A survey and outlook.
\newblock \emph{arXiv preprint arXiv:2001.04193}, 2020.

\bibitem[Ying et~al.(2021)Ying, Cai, Luo, Zheng, Ke, He, Shen, and Liu]{ying2021do}
Chengxuan Ying, Tianle Cai, Shengjie Luo, Shuxin Zheng, Guolin Ke, Di He, Yanming Shen, and Tie-Yan Liu.
\newblock Do transformers really perform badly for graph representation?
\newblock In \emph{Thirty-Fifth Conference on Neural Information Processing Systems}, 2021.

\bibitem[Yu et~al.(2022{\natexlab{a}})Yu, Wang, Vasudevan, Yeung, Seyedhosseini, and Wu]{yu2022coca}
Jiahui Yu, Zirui Wang, Vijay Vasudevan, Legg Yeung, Mojtaba Seyedhosseini, and Yonghui Wu.
\newblock Coca: Contrastive captioners are image-text foundation models.
\newblock \emph{arXiv preprint arXiv:2205.01917}, 2022{\natexlab{a}}.

\bibitem[Yu et~al.(2022{\natexlab{b}})Yu, Luo, Zhou, Si, Zhou, Wang, Feng, and Yan]{yu2022metaformer}
Weihao Yu, Mi Luo, Pan Zhou, Chenyang Si, Yichen Zhou, Xinchao Wang, Jiashi Feng, and Shuicheng Yan.
\newblock Metaformer is actually what you need for vision.
\newblock In \emph{Proceedings of the IEEE/CVF conference on computer vision and pattern recognition}, pages 10819--10829, 2022{\natexlab{b}}.

\bibitem[Yu et~al.(2022{\natexlab{c}})Yu, Tang, Rao, Huang, Zhou, and Lu]{yu2022point}
Xumin Yu, Lulu Tang, Yongming Rao, Tiejun Huang, Jie Zhou, and Jiwen Lu.
\newblock Point-bert: Pre-training 3d point cloud transformers with masked point modeling.
\newblock In \emph{Proceedings of the IEEE/CVF Conference on Computer Vision and Pattern Recognition}, pages 19313--19322, 2022{\natexlab{c}}.

\bibitem[Zeng et~al.(2022)Zeng, Attarian, Ichter, Choromanski, Wong, Welker, Tombari, Purohit, Ryoo, Sindhwani, et~al.]{zeng2022socratic}
Andy Zeng, Maria Attarian, Brian Ichter, Krzysztof Choromanski, Adrian Wong, Stefan Welker, Federico Tombari, Aveek Purohit, Michael Ryoo, Vikas Sindhwani, et~al.
\newblock Socratic models: Composing zero-shot multimodal reasoning with language.
\newblock \emph{arXiv preprint arXiv:2204.00598}, 2022.

\bibitem[Zhang et~al.(2022{\natexlab{a}})Zhang, Chen, Ouyang, Liu, Zhu, Chen, Meng, and Wu]{zhang2022pointcutmix}
Jinlai Zhang, Lyujie Chen, Bo Ouyang, Binbin Liu, Jihong Zhu, Yujin Chen, Yanmei Meng, and Danfeng Wu.
\newblock Pointcutmix: Regularization strategy for point cloud classification.
\newblock \emph{Neurocomputing}, 505:\penalty0 58--67, 2022{\natexlab{a}}.

\bibitem[Zhang and Yang(2018)]{zhang2018overview}
Yu Zhang and Qiang Yang.
\newblock An overview of multi-task learning.
\newblock \emph{National Science Review}, 5\penalty0 (1):\penalty0 30--43, 2018.

\bibitem[Zhang et~al.(2022{\natexlab{b}})Zhang, Zhao, Kang, and Shen]{zhang2022modality}
Yiyuan Zhang, Sanyuan Zhao, Yuhao Kang, and Jianbing Shen.
\newblock Modality synergy complement learning with cascaded aggregation for visible-infrared person re-identification.
\newblock In \emph{Computer Vision--ECCV 2022: 17th European Conference, Tel Aviv, Israel, October 23--27, 2022, Proceedings, Part XIV}, pages 462--479. Springer, 2022{\natexlab{b}}.

\bibitem[Zhang et~al.(2023)Zhang, Gong, Zhang, Li, Qiao, Ouyang, and Yue]{zhang2023meta}
Yiyuan Zhang, Kaixiong Gong, Kaipeng Zhang, Hongsheng Li, Yu Qiao, Wanli Ouyang, and Xiangyu Yue.
\newblock Meta-transformer: A unified framework for multimodal learning.
\newblock \emph{arXiv preprint arXiv:2307.10802}, 2023.

\bibitem[Zhou et~al.(2017{\natexlab{a}})Zhou, Lapedriza, Khosla, Oliva, and Torralba]{zhou2017places}
Bolei Zhou, Agata Lapedriza, Aditya Khosla, Aude Oliva, and Antonio Torralba.
\newblock Places: A 10 million image database for scene recognition.
\newblock \emph{IEEE transactions on pattern analysis and machine intelligence}, 40\penalty0 (6):\penalty0 1452--1464, 2017{\natexlab{a}}.

\bibitem[Zhou et~al.(2017{\natexlab{b}})Zhou, Zhao, Puig, Fidler, Barriuso, and Torralba]{zhou2017scene}
Bolei Zhou, Hang Zhao, Xavier Puig, Sanja Fidler, Adela Barriuso, and Antonio Torralba.
\newblock Scene parsing through ade20k dataset.
\newblock In \emph{Proceedings of the IEEE conference on computer vision and pattern recognition}, pages 633--641, 2017{\natexlab{b}}.

\bibitem[Zhou et~al.(2023)Zhou, Kilickaya, and Vanschoren]{zhou2023locality}
Fangqin Zhou, Mert Kilickaya, and Joaquin Vanschoren.
\newblock Locality-aware hyperspectral classification.
\newblock \emph{arXiv preprint arXiv:2309.01561}, 2023.

\bibitem[Zhou et~al.(2021)Zhou, Zhang, Peng, Zhang, Li, Xiong, and Zhang]{haoyietal-informer-2021}
Haoyi Zhou, Shanghang Zhang, Jieqi Peng, Shuai Zhang, Jianxin Li, Hui Xiong, and Wancai Zhang.
\newblock Informer: Beyond efficient transformer for long sequence time-series forecasting.
\newblock In \emph{AAAI}, 2021.

\bibitem[Zhou et~al.(2018)Zhou, Xu, and Corso]{zhou2018towards}
Luowei Zhou, Chenliang Xu, and Jason Corso.
\newblock Towards automatic learning of procedures from web instructional videos.
\newblock In \emph{Proceedings of the AAAI Conference on Artificial Intelligence}, 2018.

\bibitem[Zhu et~al.(2022)Zhu, Zhu, Li, Wu, Li, Wang, and Dai]{zhu2022uni}
Xizhou Zhu, Jinguo Zhu, Hao Li, Xiaoshi Wu, Hongsheng Li, Xiaohua Wang, and Jifeng Dai.
\newblock Uni-perceiver: Pre-training unified architecture for generic perception for zero-shot and few-shot tasks.
\newblock In \emph{Proceedings of the IEEE/CVF Conference on Computer Vision and Pattern Recognition}, pages 16804--16815, 2022.

\end{thebibliography}
}

\end{document}